\documentclass{article}

\usepackage{microtype}
\usepackage{graphicx}
\usepackage{subfigure}
\usepackage{booktabs} 

\usepackage{hyperref}

\usepackage[accepted]{icml2024}

\usepackage{amsmath}
\usepackage{amssymb}
\usepackage{mathtools}
\usepackage{amsthm}

\usepackage[capitalize,noabbrev]{cleveref}

\theoremstyle{plain}

\theoremstyle{definition}

\theoremstyle{remark}

\usepackage[textsize=tiny]{todonotes}

\begin{document}

\twocolumn[
\icmltitle{Does language matter for spoken word classification? A multilingual generative meta-learning approach}



\icmlsetsymbol{equal}{*}

\begin{icmlauthorlist}
\icmlauthor{Batsirayi Mupamhi Ziki}{equal,bytefuse}
\icmlauthor{Louise Beyers}{equal,bytefuse}
\icmlauthor{Ruan van der Merwe}{bytefuse}
\end{icmlauthorlist}

\icmlaffiliation{bytefuse}{Bytefuse}

\icmlcorrespondingauthor{Batsirayi Mupamhi Ziki}{batsi.ziki@bytefuse.ai}
\icmlcorrespondingauthor{Louise Beyers}{louise@beyers.co.za}

\icmlkeywords{spoken word classification, keyword spotting, meta-learning}

\vskip 0.3in
]

\printAffiliationsAndNotice{} 

\begin{abstract}
    Meta-learning has been shown to have better performance than supervised learning for few-shot monolingual spoken word classification. However, the meta-learning approach remains under-explored in multilingual spoken word classification. In this paper, we apply the Generative Meta-Continual Learning algorithm to spoken word classification. The generative nature of this algorithm makes it viable for use in application, and the meta-learning aspect promotes generalisation, which is crucial in a multilingual setting. We train monolingual models on English, German, French, and Catalan, a bilingual model on English and German, and a multilingual model on all four languages. We find that although the multilingual model performs best, the differences between model performance is unexpectedly low. We also find that the hours of unique data seen during training seems to be a stronger performance indicator than the number of languages included in the training data.
    
\end{abstract}

\section{Introduction}
The goal of keyword spotting is to detect pre-defined keywords in continuous audio streams \cite{deep_spoken_2022}.
One application of this is to activate voice assistants when the keywords are spoken \cite{KWS_google}.

Keyword spotting models often require datasets that contain thousands of utterances for each keyword \cite{mswc}, however, collecting data at this scale for low-resource languages is infeasible. 
As a result, it is ideal for these systems to be of a few-shot nature. 
A growing field of work shows that leveraging data from multiple languages can also improve performance on lower-resource languages \cite{artur2025, mazumder21_interspeech}. 
Mazumder et. al. \cite{mazumder21_interspeech} introduced a transfer learning method to perform few-shot keyword spotting in any language.
They did this by training a multilingual encoder via supervised learning and using it to fine-tune multilingual keyword spotting model such that it can generalise to unseen languages.
They also pointed out that the multilingual model outperformed the monolingual model on its own language and unseen languages.

Another common approach to improving the performance in few-shot systems is to use meta-learning algorithms.  
Meta-learning is informally defined as learning how to learn \cite{timothy_m_hospedales_meta-learning_2020}.
As shown in \cite{chen20j_interspeech,chen21u_interspeech}, the model-agnostic meta-learning (MAML) \cite{finn2017model} approach outperforms standard supervised learning and transfer learning in the few-shot setting.

Generative meta-continual learning (GeMCL) \cite{mohammadamin_banayeeanzade_generative_2021} is an algorithm that lies in the intersection of meta-learning and continual learning used in classification tasks.
It is a generalisation of and improvement on prototypical networks \cite{snell_prototypical}, adopting a Bayesian approach in modelling the distribution of the classes in the dataset.
Because GeMCL isolates class-specific parameters, it is immune to catastrophic forgetting \cite{mohammadamin_banayeeanzade_generative_2021}.  
Given GeMCL's strong few-shot performance and its immunity to catastrophic forgetting, it stands to be a promising algorithm to investigate further for spoken word classification settings in both the low-resource and multilingual settings.  

Our contribution is to investigate applying GeMCL's meta-learning procedure to train multilingual, bilingual and monolingual GeMCL models on the Multilingual Spoken Words Corpus (MSWC) dataset \cite{mswc}, exploring whether the multilingual model provides additional benefit over the monolingual and bilingual models on unseen languages in spoken word classification.
We show that:
\begin{enumerate}
    \item On unseen languages, the monolingual and the bilingual models' performance is comparable to the multilingual model's performance.
    \item For our setting, model performance may be more strongly linked to the number of hours of training data seen, rather than the number of languages a model is exposed to.
\end{enumerate}

\section{GeMCL}\label{sec:gemcl}

We provide an overview of the Generative meta-continual learning (GeMCL) algorithm; for more details please refer to \cite{mohammadamin_banayeeanzade_generative_2021,lee_learning_2024}.
GeMCL consists of an encoder, $f_\phi$, followed by the Bayes classifier, where each class $c$ is modelled by a Gaussian.
The encoder outputs an embedding of input $\vec{x}$ in $d$-dimensional space.
Each Gaussian has a mean, $\vec{\mu}^c \in \mathbb{R^d}$, and precision, $\vec{\lambda}^c \in \mathbb{R}_+^d$, whose estimates are updated based on the data available for that class.
The uncertainty of the mean and precision are modelled by a Normal-Gamma distribution,
\begin{equation}
    P(\vec{\mu}^c, \vec{\lambda}^c \mid \theta_0) \propto \prod_{i=1}^d \mathcal{N}\big(\mu^c_i \mid \mu_0, (\kappa_0\lambda^c_i)^{-1}\big)\mathrm{Ga}\big(\lambda^c_i \mid \alpha_0, \beta_0\big),
\end{equation}\label{eqn:gau_gemcl}
where $\theta_0 = \{\alpha_0, \beta_0, \kappa_0, \mu_0\}$ are parameters of the Normal-Gamma distribution and represent our prior belief before seeing the embedded training samples of class $c$.
However, in practice the prior of mean $\vec{\mu}^c$ is assumed to be uninformative, therefore $\kappa_0= \mu_0=0$.

The Normal-Gamma distribution is the conjugate prior to the Gaussian likelihood. Therefore, the posterior remains in the same family.
This allows closed-form updates for the Normal-Gamma parameters.

\begin{figure}[h!]
  \centering
  \includegraphics[width=\linewidth]{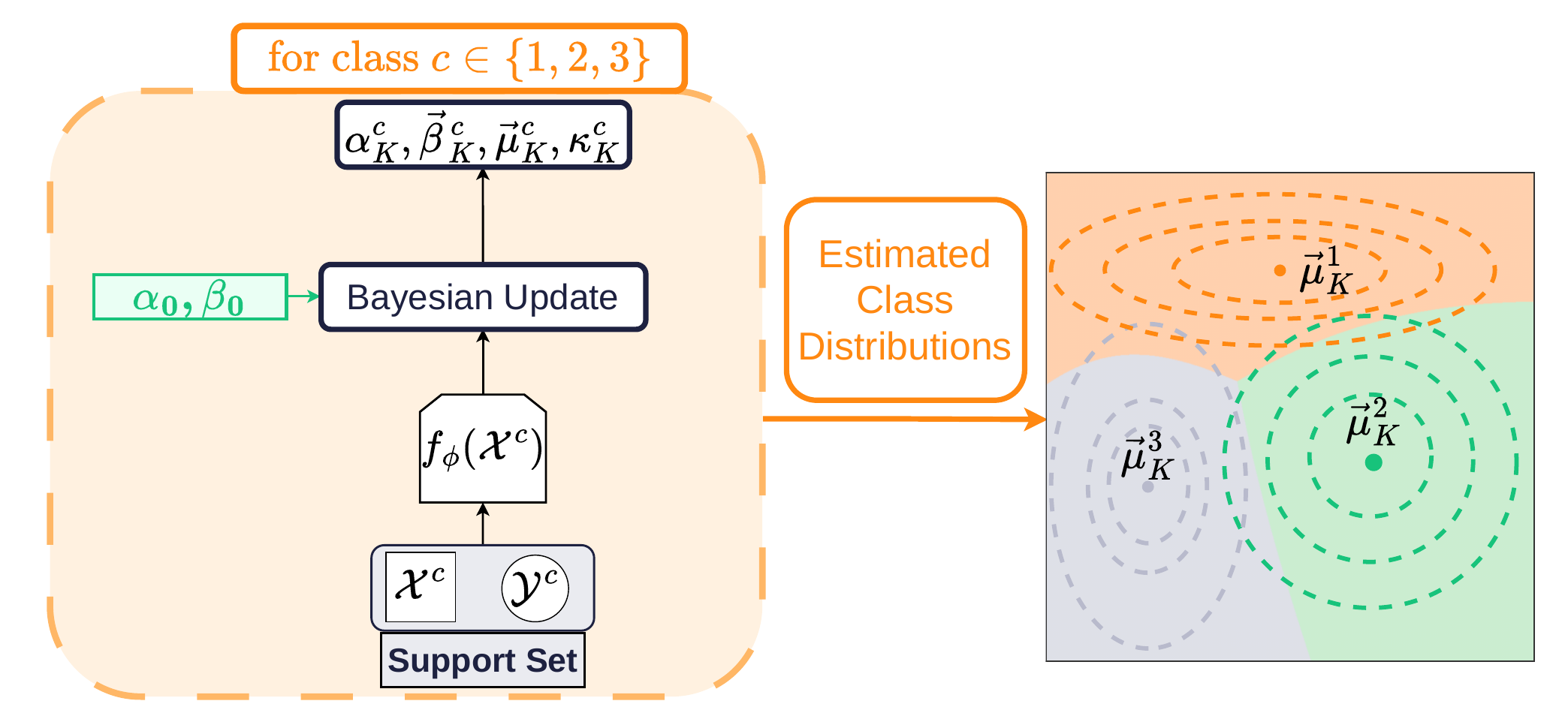}
  \caption{An example of the GeMCL procedure for learning class statistics for a classification task with three classes and $K$ samples in each class.}
  \label{fig:gemcl}
\end{figure}

Assume that each class $c$ has $K$ training samples: $(\vec{x_1},\vec{x_2}, \cdots, \vec{x_K})$.
Then the updates to the Normal-Gamma parameters are,
\begin{align}
    \kappa_K^c &= K, \label{eq:kappa} \\
    \vec{\mu}_{K}^c &= \frac{1}{K}\sum_{i=1}^K \vec{z}_i, \label{eq:mu} \\
    \alpha_K^c &= \alpha_0 + \frac{K}{2}, \label{eq:alpha} \\
    \boldsymbol{\beta}_{K}^{c} &= \beta_{0} \mathbf{1} + \frac{1}{2} \sum_{i=1}^{K} (\mathbf{z}_{i} - \boldsymbol{\mu}_{K}^{c}) \odot (\mathbf{z}_{i} - \boldsymbol{\mu}_{K}^{c} ). \label{eq:beta}
\end{align}
Here $f_\phi(\vec{x_i})=\vec{z_i}$ for $i \in \{1, 2, \cdots, K\}$ and $\odot$ denotes the Hadamard product.
The predicted label ${\hat{y}}$ is,
\begin{equation}\label{eqn:pred_student_t}
    {\hat{y}} = \underset{{y}}{\rm{argmax}}\, P\left(\vec{z} | \alpha_K^{{y}}, \vec{\beta}_K^{{y}}, \vec{\mu}_K^{{y}}, \kappa_K^{{y}} \right),
\end{equation}
where $P$ is a Student's t-distribution. Figure \ref{fig:gemcl} illustrates the GeMCL procedure for learning the class statistics.

GeMCL makes use of meta-learning to train and evaluate the encoder's parameters and the prior parameters: $\alpha_0$ and $\beta_0$.
Collectively, these parameters are known as the meta-parameters.
Meta-learning consists of two phases: meta-training and meta-testing.
A single step in meta-training involves training and testing on a batch of tasks sampled from a task distribution.
In our case, the tasks are classification tasks consisting of $N$ classes each with $K$ samples, known as a $N$-way-$K$-shot episode. 
The training samples of each class are known as the support set, while the test samples are known as the query set.
During an $N$-way-$K$-shot episode, GeMCL makes use of the support sets to learn the statistics of all the classes using equations \eqref{eq:kappa} - \eqref{eq:beta}.
Next, the model predicts the label of each sample in the query sets with equation \eqref{eqn:pred_student_t}.
The cross-entropy loss is used to obtain the gradients to update the meta-parameters.

In meta-testing, we evaluate the meta-parameters on unseen $N$-way-$K$-shot classification tasks, i.e., tasks that are not in the meta-training distribution.
This process is similar to the meta-training phase; however, the meta-parameters remain frozen.

\section{Empirical design}
In this section, we describe the MSWC dataset, the training procedure of the models, the encoder architecture of GeMCL, and the evaluation procedure.
\subsection{MSWC dataset}
We make use of the MSWC dataset \cite{mswc}.
It is an audio dataset consisting of $23.4$ million $1$-second audio clips of $340\,000$ keywords across $50$ languages.
The dataset is split into \texttt{train}/\texttt{dev}/\texttt{test} splits by the original authors for each of the keywords, prioritising speaker diversity when making the splits. We do not make use of the \texttt{dev} set.

In selecting the languages for training the multilingual models, the top four high-resource languages, English, German, French and Catalan, are used.
We filter out any words that do not have at least five valid examples. 
The words are then randomly split into the meta-training set and meta-test set with a $70:30$ split. Table \ref{tab:meta_splits} shows the meta-splits for the four training languages.

\begin{table}[h!]
  \centering
  \caption{The number of meta-training and meta-test classes for the four training languages and the multilingual training procedure.}
  \label{tab:meta_splits}
  \begin{tabular}{l r r } 
    \toprule
    \textbf{Language} & \textbf{Meta-train classes} & \textbf{Meta-test classes} \\
    \midrule
    English & $8\,915$ & $3\,821$ \\ 
    German  & $6\,358$ & $2\,726$ \\ 
    French  & $4\,742$ & $2\,033$ \\ 
    Catalan & $4\,608$ & $1\,975$ \\ 
    \bottomrule
  \end{tabular}
\end{table}

The other languages, i.e., the unseen languages, undergo a similar filtering system, additionally disregarding any language that does not have enough data for a $25$-way-$5$-shot classification task.
The unseen languages are only used for evaluation hence all their valid words are used to evaluate the models.
This gives $35$ unseen languages, totalling $39$ languages when combined with the training languages.

\subsection{Monolingual and multilingual training} 
There are four monolingual models: English, German, French and Catalan, which are trained on the meta-training classes of their respective languages.
The bilingual model is trained on English and German, whereas the multilingual model is trained on all four languages.
For all models, $16$ batches of $25$-way-$5$-shot episodes are used at each step in the meta-training procedure.

For each episode, we randomly select $25$ words from the meta-training classes.
Next, we randomly select five recordings from the \texttt{train} split of each word to form the support set for that word in the current episode.
Similarly, we randomly select five recordings from the \texttt{test} split of each word to form the query set for the current episode.
We then perform meta-training as described in Section \ref{sec:gemcl}.

The multilingual model's batch of $16$ episodes consists of four episodes from each of the training languages, whereas the bilingual model's batch consists of eight episodes each from German and English.

\subsection{GeMCL implementation details}
Our implementation of GeMCL is inspired by \cite{lee_learning_2024}.
The encoder is a  $12$-layer-$12$-head transformer with $85\,066\,756$ parameters.
The encoder takes mel-frequency cepstral coefficients (MFCCs) as input.
MFCCs are extracted from waveforms sampled at $16$ kHz, using a frame length of $25$ ms, a frame shift of $10$ ms, and $40$ mel filterbanks.
We retain the first $13$ cepstral coefficients.
To optimise the meta-parameters, we employ the AdamW \cite{adamw} optimiser with a weight decay of 1e-2 and a learning rate of 5e-5.
The models are meta-trained for $2\,000$ steps.

\subsection{Evaluation}
We evaluate the models on all $39$ languages.
For the four training languages the models are evaluated only on the reserved meta-test classes, which ensures that all models are evaluated on the same hold-out set.

We use $100$ episodes per language to evaluate each model.
Each episode is $25$-way-$5$-shot.
To assess whether the differences in accuracy between the multilingual model and each of the other models are statistically significant, we use the bootstrap percentile method.
For each language, we compute a $95\%$ confidence interval on the difference in mean accuracy between each model and the multilingual model using $9\,999$ resamples
For each of $9\,999$ resamples, we independently sample with replacement $100$ episode accuracy values from each model and record the difference of their means.
If the confidence interval of the difference excludes zero, we conclude that the difference is statistically significant; if it includes zero, the difference is not statistically significant \cite{ferrer2024goodpracticesevaluationmachine}.

\section{Results \& Discussion}
Table \ref{tab:seen_lang_results} reports the average accuracy of the multilingual model and the monolingual models on their respective training languages.
The monolingual models produce marginally higher accuracy on their respective languages.
However, the differences in accuracies are not statistically significant, suggesting that the representations produced by the multilingual model provide no meaningful benefit over those produced by the monolingual models.
This agrees with the suggestion of \cite{zhu-etal-2024-taste} that for high resource languages, multilingual models do not provide any additional benefit over monolingual models.

\begin{table}[h]
  \centering
  \caption{Mean accuracy over 100 episodes (25-way-5-shot) on seen languages, 
  comparing each language's corresponding monolingual model (e.g., for Catalan, 
  this is the Catalan monolingual model) and 
  the multilingual model.}
  \label{tab:seen_lang_results}
  \begin{tabular}{lrr}
    \toprule
    \textbf{Language} & \textbf{Monolingual (\%)} & \textbf{Multilingual (\%)} \\
    \midrule
    English & $88.78$ & $88.10$   \\
    German  & $93.99$ & $93.96$   \\
    French  & $89.46$ & $89.17$  \\
    Catalan & $93.57$ & $92.95$ \\
    \bottomrule
  \end{tabular}
\end{table}

The multilingual model performs significantly better than the bilingual model in 11 languages.
Compared to the monolingual models, it shows significantly better performance in 29 (English), 28 (German), 35 (French), and 38 (Catalan) languages.
It should be noted that for every language in which a monolingual model achieved higher accuracy than the multilingual model, the difference in per-language accuracy between the two models was not statistically significant.
The only language for which the bilingual model performs significantly better than the multilingual model is Hakha Chin.

\begin{figure}[h!]
  \centering
  \includegraphics[width=\linewidth]{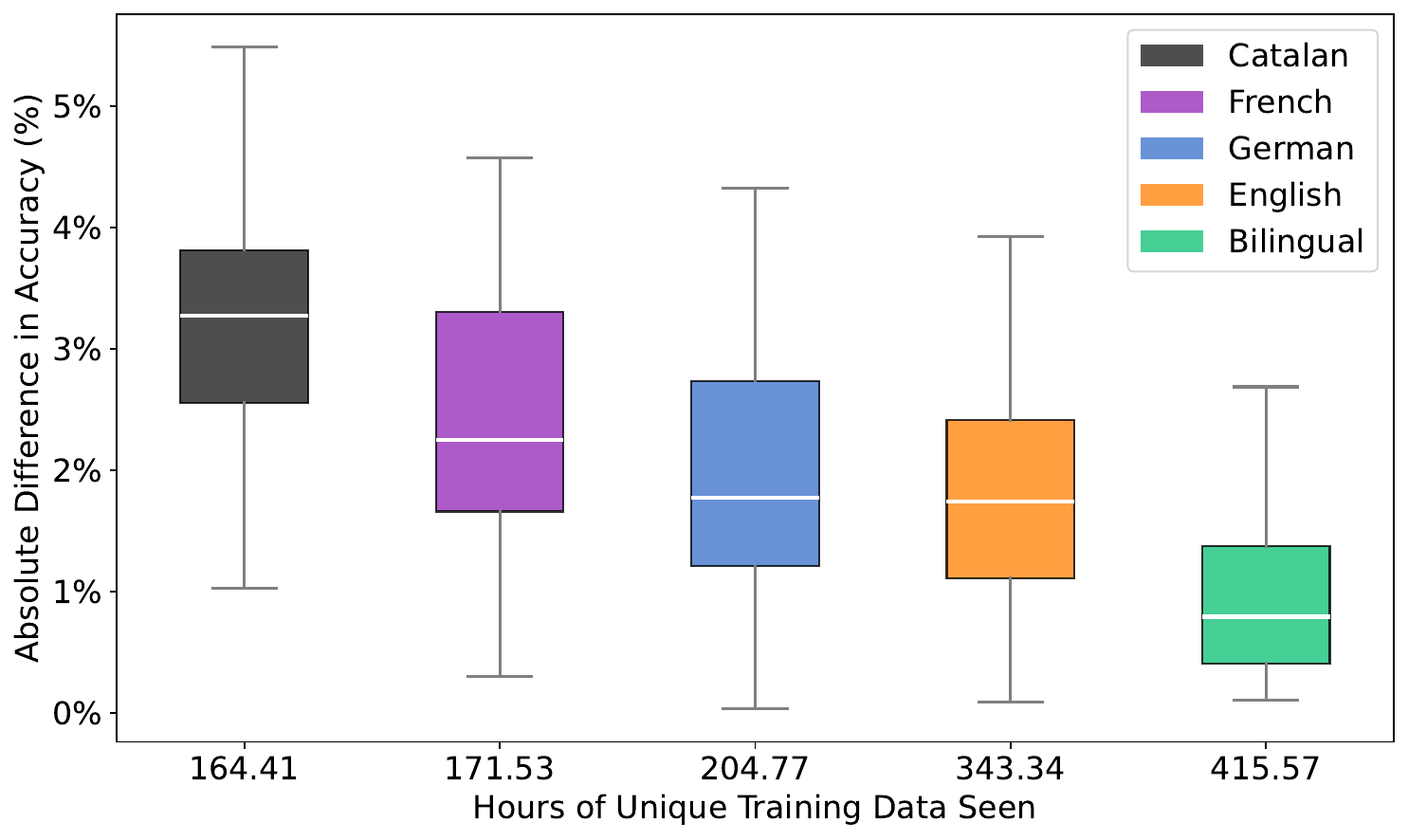}
  \caption{Box plots showing the absolute difference in mean accuracy between each model and the 4-language multilingual reference model, across all 39 evaluation languages. Each box summarises the distribution of per-language absolute differences. Models are ordered by hours of unique training data seen.}
  \label{fig:box_and_whiskers}
\end{figure}

Figure \ref{fig:box_and_whiskers} illustrates the box plot of the absolute difference in the mean accuracy between each model and the multilingual model.
Considering that recordings are reselected for each batch, the total unique recordings seen by GeMCL may differ from training run to training run. We simulate our sampling strategy for the duration of training, and find that over $10$ simulations the total duration of unique recordings seen averages to the hours reported by Figure \ref{fig:box_and_whiskers}.

As the hours of unique training data seen increases, we see that the average absolute difference in accuracy decreases, with the bilingual model being the closest to the multilingual model.

Despite the fact that the multilingual model was trained on two additional languages compared to the bilingual model from Figure \ref{fig:box_and_whiskers}, we see that the average absolute difference in accuracy between the bilingual model and the multilingual model is roughly $1\%$.
Since the bilingual and multilingual models produce such similar performance, we exclude the bilingual model from our per-language analysis.

\begin{figure}[h!]
  \centering
  \includegraphics[width=\linewidth]{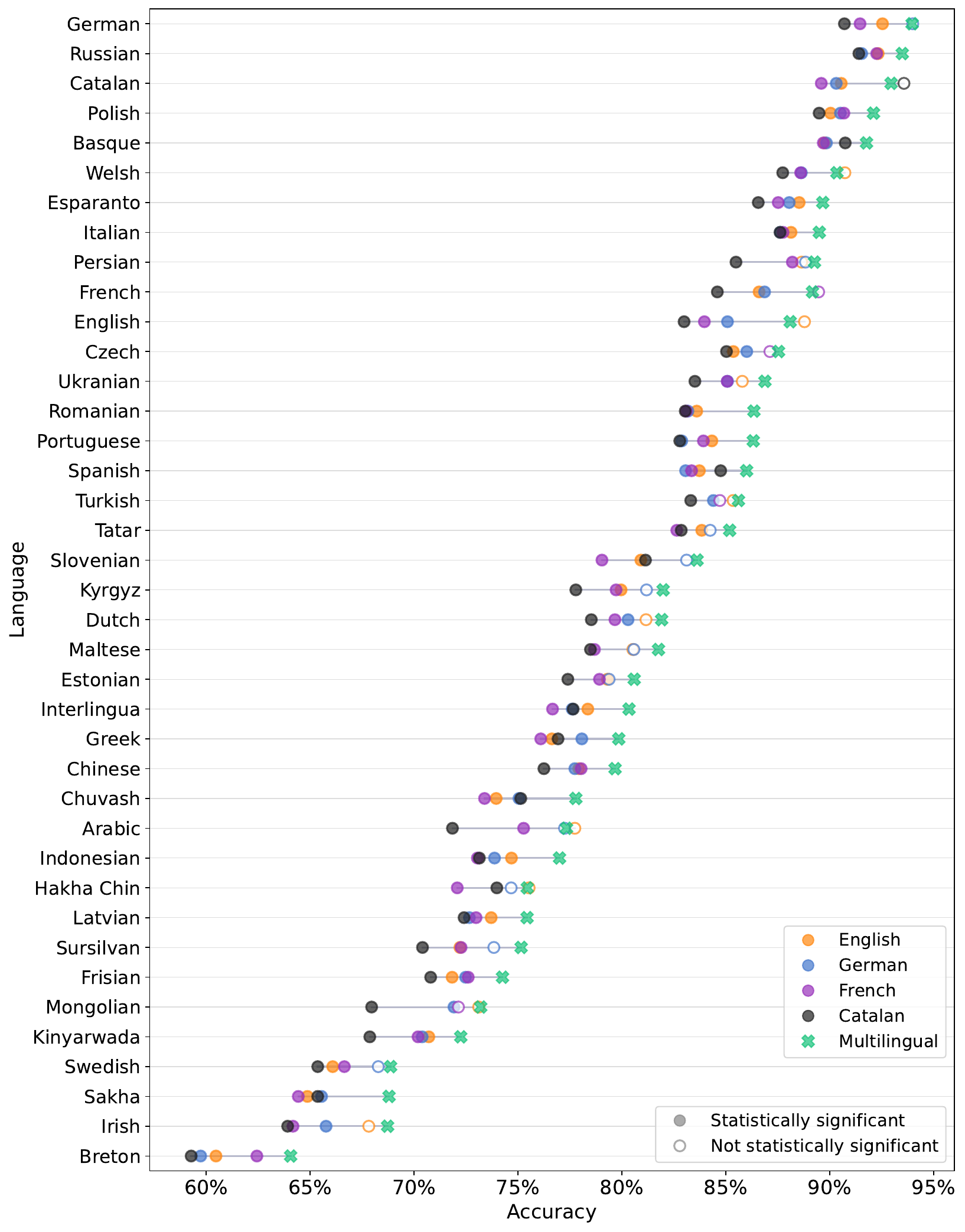}
  \caption{Per-language accuracy of four monolingual models and a multilingual model trained on all four languages, evaluated on 25-way 5-shot spoken word classification across 39 languages. Each circle represents a monolingual baseline. Languages are sorted by multilingual accuracy in descending order. A filled circle indicates that the difference between the monolingual and multilingual model is statistically significant. A hollow circle indicates no statistically significant difference.}
  \label{fig:final_results}
\end{figure}

Figure \ref{fig:final_results} illustrates the performance of the monolingual and multilingual models on all 39 languages.
The difference in per-language average accuracy between a monolingual model and the multilingual model is never more than $6\%$.
For most of the languages, the monolingual models, particularly English and German, perform comparably to the multilingual model.

We have found that one can meta-learn on data from one high resource language from the MSWC dataset and achieve performance comparable to training on many languages from MSWC. While it is unknown whether this holds across more meta-learning paradigms, languages, and datasets, the difference in performance is remarkably small (note the scale of Figure \ref{fig:final_results}). 

Further note the differences between the monolingual models. Both Figures \ref{fig:final_results} and \ref{fig:box_and_whiskers} show a somewhat consistent ordering of the performances of the monolingual models. Additionally, the order of the monolingual model performance appears to follow the order of the number of hours of data on which each model was trained. It is therefore reasonable to suspect that the number of hours of training data significantly influences the performance of the model even at the scale of the data used in our experiments, though further research is required. If we assume that it is the case, though, then since the multilingual model was trained on far more data ($426.54$ hours) than the monolingual models (German, French and Catalan were trained on less than half that), it remains unclear whether the multilingual model's performance can be attributed to the variety of languages it saw, or whether the volume and word diversity of its data is a larger factor in its success.

It is also of note that the order and spread of the average accuracies of the monolingual models can be sensitive to the language on which they are evaluated. Further investigation would be required to determine which factors might contribute to this sensitivity.

\section{Conclusion}
We applied the GeMCL meta-learning framework to spoken word classification using the MSCW dataset, training 
multilingual, bilingual and monolingual models.
The multilingual model was trained on the four most resourced languages: English, German, French, and Catalan. We also trained four monolingual models, one for each of these languages, and a bilingual model trained on English and German.
When comparing the monolingual models to the multilingual model on their respective training language we found the multilingual model does not provide any additional benefit as the performances of each is nearly identical.
On all of the languages, the bilingual model's average absolute difference in accuracy relative to the multilingual model was $1\%$, whereas the monolingual models underperformed the multilingual model by only a small margin, with the English-only model achieving within $2\%$ of the multilingual model on average. We therefore found that the benefit of adding languages to the training data is lower than originally expected when using GeMCL for multilingual spoken word classification. The results suggests that the volume of data that the model sees has an effect, since the performance of the monolingual models could be ordered by the hours of data they were trained on. Further investigation is required to determine the significance of this finding.

\section{Generative AI use disclosure}
Generative AI assistance was used in producing the code for experiments of this paper, as well as for providing a starting point for some of the figures which appear in this paper. Generative AI assistance was not used to write this paper.

\bibliography{bib}
\bibliographystyle{icml2024}

\end{document}